\documentclass{article}

\usepackage{PRIMEarxiv}
\usepackage{wrapfig} %
\usepackage{caption} 
\usepackage[utf8]{inputenc} 
\usepackage[T1]{fontenc}    
\usepackage{hyperref}       
\usepackage{url}            
\usepackage{booktabs}       
\usepackage{amsfonts}       
\usepackage{nicefrac}       
\usepackage{microtype}      
\usepackage{lipsum}
\usepackage{fancyhdr}       
\usepackage{graphicx}       
\usepackage{longtable}
\usepackage{multirow}
\usepackage{amssymb}
\graphicspath{{media/}}     

\title{3CEL: a Corpus of Legal Spanish Contract Clauses
}

\author{
  Nuria Aldama García\\
  Instituto de Ingeniería del Conocimiento \\
\texttt{nuria.aldama@iic.uam.es} \\
  \And
  Patricia Marsà Morales\\
  Instituto de Ingeniería del Conocimiento \\
\texttt{patricia.marsa@iic.uam.es} \\
  \And
  David Betancur Sánchez\\
  Instituto de Ingeniería del Conocimiento \\
\texttt{david.betancur@iic.uam.es} \\
  \And
  Álvaro Barbero Jiménez\\
  Instituto de Ingeniería del Conocimiento \\
\texttt{alvaro.barbero@iic.uam.es} \\
  \And
  Marta Guerrero Nieto\\
  Instituto de Ingeniería del Conocimiento \\
\texttt{marta.guerrero@iic.uam.es} \\
 \And
  Pablo Haya Coll\\
  Instituto de Ingeniería del Conocimiento \\
\texttt{pablo.haya@iic.uam.es} \\
  \And
  Patricia Martín Chozas\\
  Ontology Engineering Group \\
  Universidad Politécnica de Madrid \\
\texttt{pmchozas@fi.upm.es} \\
\And
  Elena Montiel Ponsoda\\
  Ontology Engineering Group \\
  Universidad Politécnica de Madrid \\
\texttt{elena.montiel@upm.es} 
\\
}

\begin{document}
\maketitle

\begin{abstract}
Legal corpora for Natural Language Processing (NLP) are valuable and scarce resources in languages like Spanish due to two main reasons: data accessibility and legal expert knowledge availability. INESData 2024 is a European Union funded project lead by the Universidad Politécnica de Madrid (UPM) and developed by Instituto de Ingeniería del Conocimiento (IIC) to create a series of state-of-the-art NLP resources applied to the legal/administrative domain in Spanish. The goal of this paper is to present the Corpus of Legal Spanish Contract Clauses (3CEL), which is a contract information extraction corpus developed within the framework of INESData 2024. 3CEL contains 373 manually annotated tenders using 19 defined categories (4 782 total tags) that identify key information for contract understanding and reviewing. 
\end{abstract}

\keywords{Corpus linguistics \and Information extraction \and Spanish legal domain \and Span categorization}

\section{Introduction}

Information extraction (IE) is defined as the NLP task that deals with the identification of particular pieces of information in unstructured documents \cite{grishman97,piskorskiyangarber12,dagdelenetal24}. In other words, the main objective of IE is to spot predefined relevant information in raw text. IE includes different subtypes depending on the nature of the information to be extracted. Thus, Named Entity Recognition (NER), Co-Reference Resolution, Relation Extraction or Event Extraction are encompassed under the umbrella of IE \cite{piskorskiyangarber12}. 

IE encounters specific challenges, particularly with regard to data availability and the need for expert knowledge. First, access to raw data is limited depending on the target domain (e.g. health, law, insurance...). Second, achieving proper descriptions that account for the information to be extracted is challenging and it is important to count on expert knowledge to avoid ambiguity. Besides, background knowledge is essential to separate relevant information from spurious or additional details. Specialists willing to participate in validation and resource creation procedures are not always available and usually have limited time. Finally, creating quality and manually annotated corpora is time consuming and expensive: creating annotation guides based on expert knowledge, training annotators, applying blind and peer annotation methodologies or quality metrics implies investing time and economic resources. These limitations make the availability of high quality IE resources scarce. This lack of resources is particularly noticeable in Spanish. For example, to date there are 423 public token classification datasets in English in HuggingFace, but only 88 in Spanish\footnote{Search date: November 19$^{th}$, 2024. Search link: \url{https://huggingface.co/datasets?task_categories=task_categories:token-classification&language=language:es&sort=trending$}}.

The objective of this paper is to present one of the state-of-the-art NLP resources applied to the legal domain in Spanish,
namely, the Corpus of Legal Spanish Contract Clauses (3CEL). To the best of our knowledge, no corpus for legal clause extraction comparable to 3CEL is currently available in Spanish. 3CEL is shared in the INESData Legal Data Space \footnote{3CEL link: to be published}. INESData is a project funded by the Spanish Ministry for Digital Transformation and Civil Service and the EU  (NextGenerationEU). In the context of INESData, which is lead by Universidad Politécnica de Madrid (UPM), the Instituto de Ingeniería del Conocimiento (IIC) is working towards the development of a Legal Data Space demonstrator in a distributed cloud infrastructure. Data Spaces have been defined as secure and interoperable ecosystems that enable data providers, intermediaries, and consumers to share data and services. 

This paper is structured as follows. Section 2 includes a brief review of previous work in corpus creation and modeling for legal information extraction. Section 3 describes the creation process of 3CEL. Section 4 depicts the experiments conducted to demonstrate that 3CEL is a valuable resource for model fine-tuning. Section 5 concludes the paper and defines future lines of work. 

\section{Previous work in Legal NLP}
The state of the art on legal NLP encompasses tools and resources to be used in different languages and tasks. Among those resources, ontologies that define and structure legal key concepts have evolved over time. LegalRuleML \cite{Legalruleml21}, Financial Industry Business Ontology - FIBO \cite{fibo99}, Contract Ontology \cite{contractonto20}, Event Extraction in Labour Law \cite{martinchozas21} or Public Procurement Ontology – PROOC \cite{prooc16} present different approaches to standardize legal information annotations to provide common vocabularies for finance, contracts and related concepts.   

European Union funded platforms like They Buy for You \cite{theybuyforyouxx} or NextProcurement \cite{nextprocurement} comprise legal documents and metadata databases that can be explored by means of a series of online tools (e.g. knowledge graphs). Their goal is to ease access, improve information quality and look for standardization so anyone can make use of legal information. Another project worth mentioning is the EU-funded innovation action Lynx \cite{montiel17} \cite{martinchozas19}. Lynx developed a knowledge-based AI service platform to process, enrich and analyze legal documents. The focus of the platform was to assist companies in addressing compliance issues in a multilingual and multi-jurisdictional scenario. The platform relied on a data model to structure and link documents in a Legal Knowledge Graph (LKG), and on a set of NLP and Information Retrieval (IR) services to process legal documents.

\begin{figure}[h]
\centering
\includegraphics[width=0.90\textwidth]{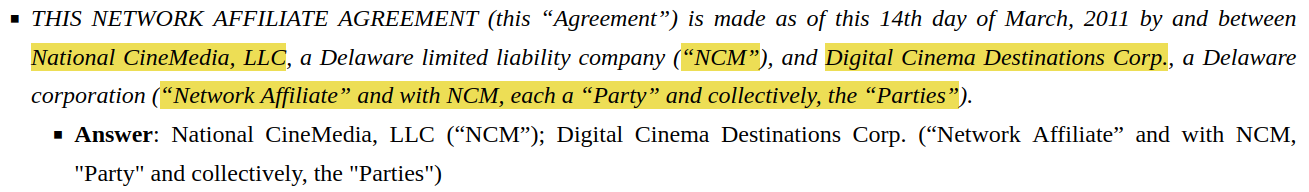}
    \caption{CUAD annotation example for the label '\textit{parties}'.}
    \label{cuad_parties}
\end{figure}

In the United States, the Atticus Project platform \cite{atticus23} develops corpora and training courses to explore the interfaces between the legal domain and artificial intelligence. One of their reference resources is the Contract Understanding Atticus Dataset (CUAD) \cite{cuad21}. CUAD contains 510 commercial legal contracts from United States and more than 13\,000 manually annotated labels using a 41-label tag set that defines types of legal clauses that are considered important by experts in contract reviewing. CUAD is annotated assigning a label to each text segment of interest\footnote{Text segments are generally complete sentences, although certain labels apply to shorter fragments of text like ‘parties’ or ‘agreement date/effective date’.}. CUAD annotation guidelines propose to transcribe the actual text that is classified under certain target labels to tackle question answering tasks (see in Figure \ref{cuad_parties}). CUAD is used to fine-tune BERT \cite{BERT}, ALBERT \cite{ALBERT}, RoBERTa \cite{ROBERTA} and DeBERTa \cite{DEBERTA} models. DeBERTa attains an area under the precision recall curve (AUPR) of 47.8\,\%, a precision of 80\,\%, and a recall of 44.0\,\%. According to the authors, these results are already a promising approach to save a substantial amount of time to lawyers when compared to reading an entire contract, although there is still room for improvement due to the task complexity. Since the essence and main objective of CUAD and 3CEL are aligned, CUAD has been a key resource to the design of 3CEL annotation scheme.

Regarding Spanish legal-domain models,  it is important to mention RoBERTalex \cite{gutierrezfandino2021legal} or \texttt{legal-xlm-roberta-large} \cite{niklaus23}. These models are trained to solve the next token prediction task and be further fine-tuned to perform any other Natural Language Understanding (NLU) tasks. Additionally, \texttt{littlejohn-ai/bge-m3-spa-law-qa} \cite{littlejohn24} is fine-tuned to calculate sentence similarity in legal documents. However, to the best of our knowledge, there is no Spanish legal domain span categorization (see 4.1) model currently available. 

\section{Corpus creation}
\begin{figure}[h]
\centering
\includegraphics[width=0.9\textwidth]{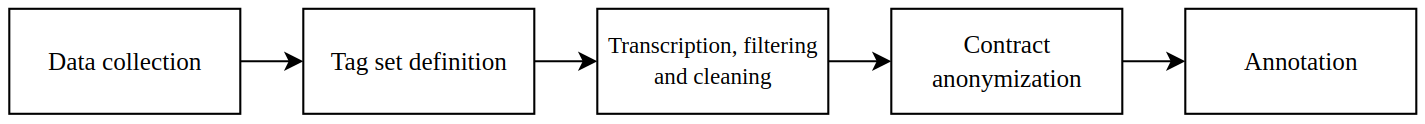}
    \caption{Corpus creation methodology.}
    \label{corpus_creation_steps}
\end{figure}

3CEL is developed in five steps following the IIC methodology on corpus creation \cite{aldamaetal22}: data collection, tag set definition, document transcription, filtering and cleaning, anonymization and annotation (see Figure \ref{corpus_creation_steps}). The following subsections dive deep into the technical aspects of each of these five processes. 

\begin{figure}[h]
\centering
\includegraphics[width=0.9\textwidth]{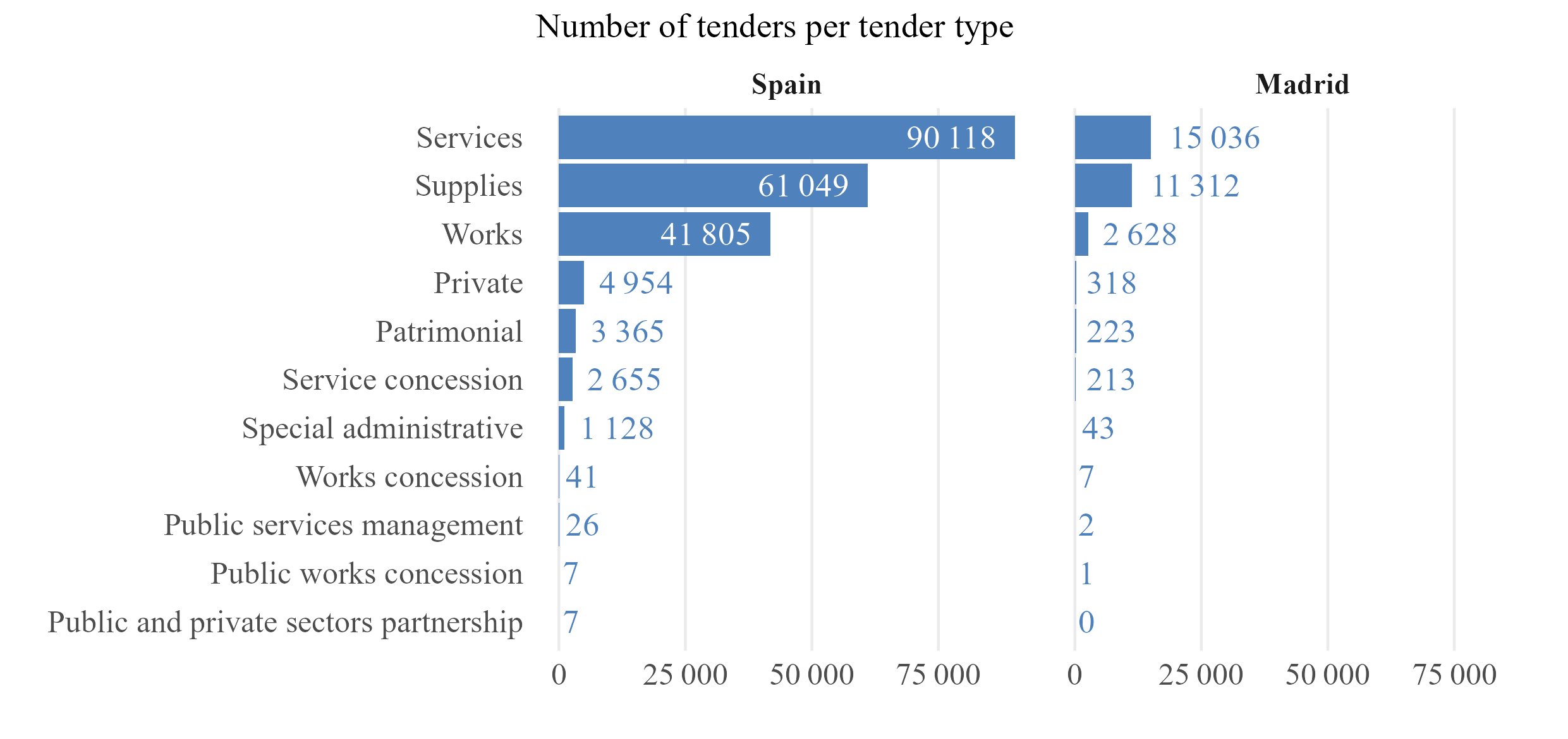}
    \caption{Spain and Madrid tender typology distributions.}
    \label{spain_madrid_tender_typology}
\end{figure}

\begin{figure}[h]
\centering
\includegraphics[width=0.9\textwidth]{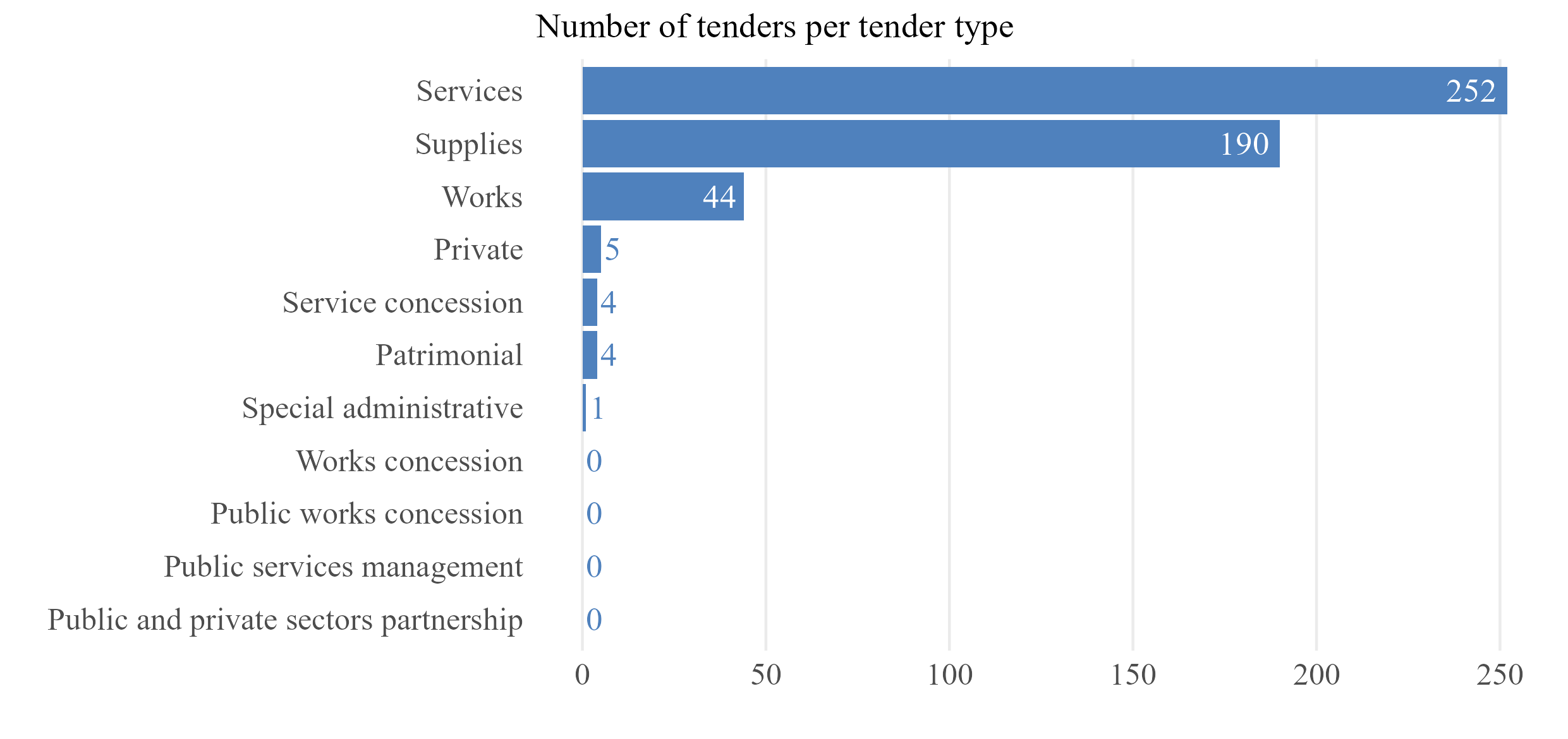}
    \caption{500-sample tender typology distribution.}
    \label{sample_tender_typology}
\end{figure}

\textbf{Data collection.} 3CEL contains tenders published by the Spanish Public Sector Procurement Platform \cite{contratacionpublica08}. Tenders are written in Spanish and were executed in the region of Madrid between December 2021 and December 2023. The Public Sector Procurement Platform provides a contract typology used to extract the distributions in Figures \ref{spain_madrid_tender_typology} and \ref{sample_tender_typology}. All contracts are downloaded in PDF format. The remaining of this section explains how Spanish Public Procurement data was filtered to obtain the sample to build the 3CEL.

\begin{wraptable}{r}{8cm}
  \centering
  \begin{tabular}{lp{4.5cm}}
    \toprule
    \textbf{Parameter}     & \textbf{Value}                                             \\
    \midrule
    Country                & Spain (\textit{España})                                            \\
    Execution place        & ES30 Madrid Region (\textit{Comunidad de Madrid})                                   \\
    Contract type          & \textit{The one according to the distribution}\\
    State                  & Resolved (\textit{Resuelta})                                                   \\
    Submission date        & December 20th, 2021 – December 20th, 2023                  \\
    \bottomrule \\ 
  \end{tabular}
    \caption{Data filtering parameters.}
    \label{tab:data_filtering_parameters}
\end{wraptable}

Figure \ref{spain_madrid_tender_typology} shows the distribution of contract typology provided by the Spanish Sector Procurement Platform for Spain and Madrid regions. According to the Spanish Sector Procurement Platform classification, there are 11 contract types. The three most represented types in Spain and Madrid during the selection period are Supplies (\textit{Suministros}), Services (\textit{Servicios}) and Works (\textit{Obras}). These three categories represent more than 95\,\% of the documents for both Spain and the region of Madrid. The least represented categories are Public utilities/services management (\textit{gestión de servicios públicos}), Public works concession (\textit{concesión de obras públicas}) and public-private partnership (\textit{colaboración entre el sector público y el sector privado}), corresponding to less than the 0.03\,\% of the sample.  

\begin{figure}[h]
\centering
\includegraphics[width=0.9\textwidth]{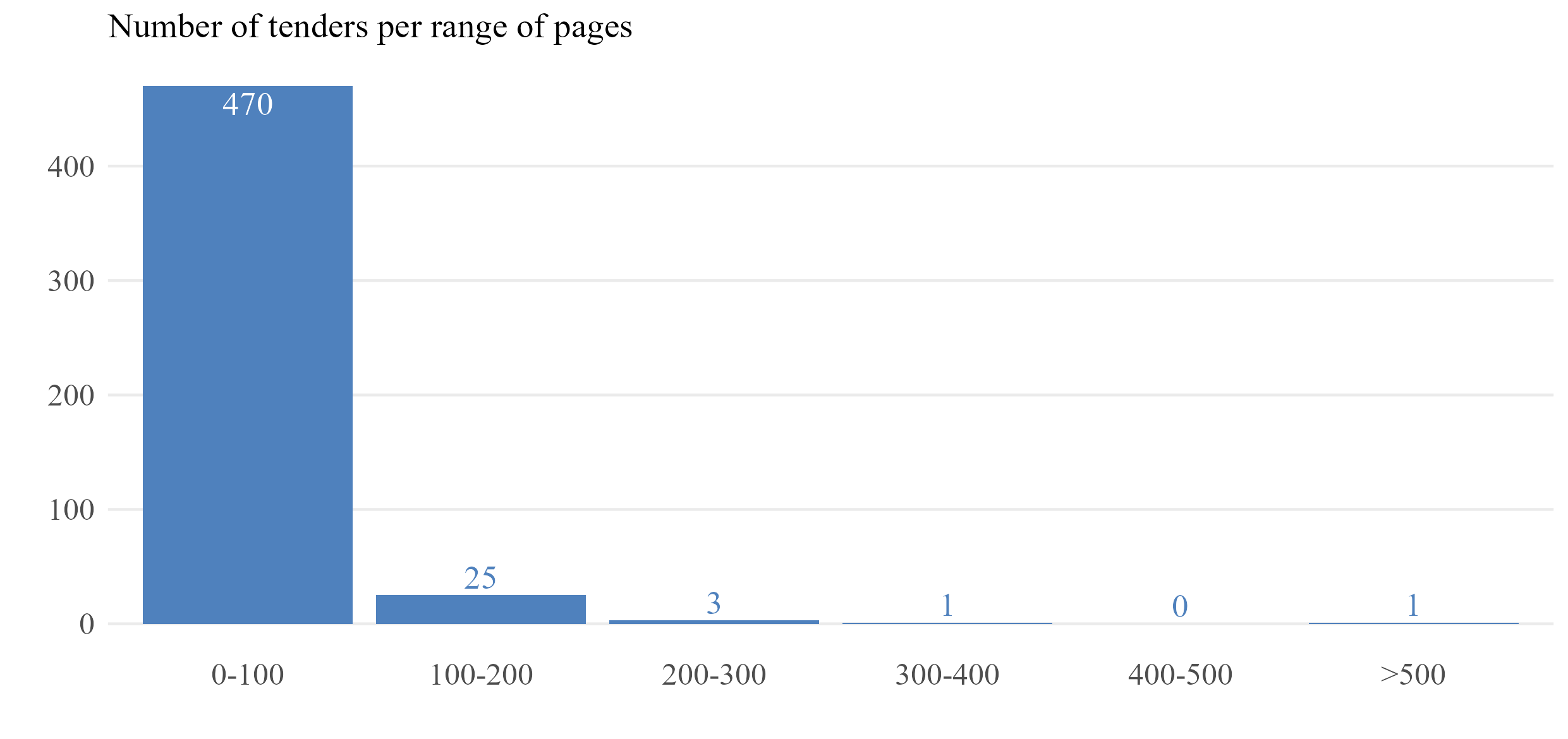}
    \caption{Number of tenders per range of pages distribution.}
    \label{page_length_distribution}
\end{figure}

Due to annotation time limitations, the sample used as the basis to build 3CEL contains 500 contracts whose distribution is calculated taking as a reference the distribution for Madrid presented in Figure \ref{spain_madrid_tender_typology}, for the sake of consistency. To gather the 500 contracts, the search engine provided by the Spanish Public Sector Procurement Platform is used applying the filtering parameters in Table \ref{tab:data_filtering_parameters}. Figures \ref{sample_tender_typology} and \ref{page_length_distribution} show the contract typology and length distribution of the 500-tender sample, respectively.  

\textbf{Tag set definition.} Category selection and definition in legal information extraction is a crucial step to ensure that the following annotation phase adequately captures the range of relevant legal information. Intuitively, the idea is to list the most useful and interesting pieces of information to be extracted from contracts to ease access to key information in processes like contract reviewing or user understanding. This general objective is formalized as follows. 

CUAD \cite{cuad21} serves as a starting point for tag selection. However, due to jurisdictional differences between Spanish and U.S. legal systems and the differences in the nature of the contracts (private commercial contracts vs tenders), a new tag set is proposed (see Table \ref{tab:3CEL_label_set}). Thus, a lawyer expert in contract reviewing set 72 potentially relevant pieces of information or tags to be annotated. Since 72 tags exceeds the scope of the project in terms of annotation time, three criteria were applied to determine the final tag set: first, the degree of relevance a piece of information has for a contract expert. Second, the possibility of finding the selected pieces of information in both public and private contracts. Third, categories must have a high potential representation in the data through key word searches. Underrepresented categories were ruled out from the tag set. The final tag set contains 19 labels related to contract provisions, contract crisis and compliance issues. Once the tag set is formed, each of the labels is further defined and applied to real contract contexts. Tag definitions and application examples are the cornerstones to build the 3CEL annotation guidelines. 

\begin{table}
    \begin{center}
    \begin{tabular}{ll}
        \toprule
        \textbf{Contract field} & \textbf{Label}\\
        \midrule
        \multirow{11}{7em}{Contract provisions}
        & Object (\textit{objeto})\\ 
        & Contract price (\textit{precio del contrato})\\ 
        & Canon (\textit{canon})\\
        & Contract duration (\textit{duración del contrato})\\ 
        & Exclusivity (\textit{exclusividad})\\ 
        & Verification of contract compliance (\textit{verificación del cumplimiento})\\
        & Warranty duration (\textit{plazo de garantía})\\ 
        & Guarantee (\textit{garantía económica})\\ 
        & Bank guarantee (\textit{garantía - aval bancario})\\
        & Surety board (\textit{garantía - seguro de caución})\\ 
        & Price retention (\textit{garantía - retención de precio})\\ 
        \midrule
        \multirow{5}{7em}{Contract crisis} 
        & Termination (\textit{resolución})\\ 
        & Delay penalty (\textit{penalidad por demora})\\ 
        & Defective performance penalty (\textit{penalidad por cumplimiento defectuoso})\\
        & Damage award (\textit{indemización de daños y perjuicios por incumplimiento})\\ 
        & Third-party damage compensation (\textit{indemnización de daños a terceros})\\
        \midrule
        \multirow{3}{7em}{Compliance} & Personal data protection (\textit{protección de datos personales})\\ 
        & Intellectual property - IP (\textit{propiedad intelectual})\\ 
        & Confidentiality (\textit{confidencialidad})\\
        \bottomrule
    \end{tabular}
    \end{center}
        \caption{3CEL tag set.}
        \label{tab:3CEL_label_set}
\end{table}

\textbf{Transcription, filtering and cleaning.} Three processing steps are needed to convert source documents into a suitable format: PDF transcription, filtering data and cleaning data. First, PDF conversion to raw text is needed, in order for the 3CEL to be annotated. Second, a filtering process is applied to tenders to ensure data relevance. Last, a two phase-cleaning process is carried out to avoid transcription noise from the texts. 

To facilitate accurate text extraction, OCRmyPDF \cite{ocrmypdf23}, PDFTOTEXT \cite{pdftotext21}, PDFPLUMBER \cite{pdfplumber24} and PYPDF2 \cite{pypdf222} are tested to identify the library that works and generalizes best in this case. To test the tools, a random sample of 24 tenders is processed and manually peer reviewed. After testing, the selected transcription tool is OCRmyPDF. 458 out of 500 (91.6\,\%) tenders are transcribed. The remaining 8.4\,\% are ruled out due to total or partial missing information. 

Once the transcribed tenders are ready, an additional filtering process is carried out to ensure data diversity and relevance.  To begin with, contracts exceeding 40 pages (see Figure \ref{page_length_distribution}) and containing fewer than 3 categories\footnote{Categories are searched using lists of key words as a means of preliminary categorizing information and obtaining statistics.} are removed. Besides, texts are further filtered by means of contract type, so the dataset complies with the original contract typology distribution. 

Cleaning is carried out in two steps: an automated phase and a manual one. In the automated cleaning phase, the text is refined to correct formatting inconsistencies and ensure a more uniform structure. This process addresses issues related to spacing, special characters and overall text presentation. Following the automated phase, a manual revision is performed on the remaining contracts. 373 out of the 458 contracts are retained after this manual cleaning. During this phase, headers, footers and texts in the margins of contracts are addressed and manually removed. Additionally, line breaks and spaces that were not automatically cleaned are replaced with the appropriate line breaks. 

\textbf{Contract anonymization.} Anonymizing sensitive information in legal texts is a critical task to ensure compliance with privacy regulations. An anonymization process typically involves identification and replacement of different types of personally identifiable information (PII), such as names or ID numbers. In this project, a comprehensive anonymization approach is implemented to handle the entities presented in Table \ref{tab:anonymization_entities} regarding Spanish legal/administrative domain, focusing on two different techniques: a system based on a NER model and a pattern-matching strategy using regular expressions. 

Regarding person and location entity detection, the pre-trained model \texttt{MMG/xlm-roberta-large-ner-spanish} is used \cite{mmg23}. Once person or location entities are identified, they are replaced with realistic but fake person or location names from different lists of linguistic resources created for the sake of this project. The way person and location entities are anonymized in 3CEL ensures consistency and coherence of the contexts where entities appear. Given a detected entity, it is always replaced using the same string, that is, detected entities and anonymized ones are unambiguously related to preserve the contexts where they appear. Both contextual coherence and consistency have a positive impact on model fine-tuning.

\begin{wraptable}{r}{8cm}
  \centering
  \begin{tabular}{p{3cm}p{4cm}}
    \toprule
    \textbf{Detection strategy}    & \textbf{Entity type}    \\
    \midrule
    NER information                & Person, location (postal address) \\
    Matched-pattern information & ID, Telephone number, Email, Bank account number, Health card number
    \\
    \bottomrule \\ 
  \end{tabular}
    \caption{Summary of anonymization target entities.}
    \label{tab:anonymization_entities}
\end{wraptable}

\begin{figure}[h]
\centering
\includegraphics[width=0.9\textwidth]{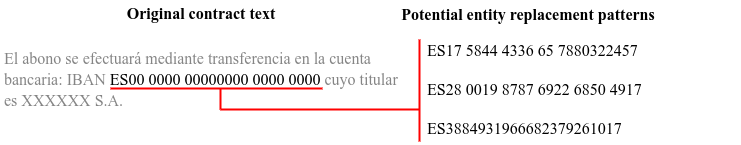}
    \caption{Random bank account number pattern generation.}
    \label{iban_substitution}
\end{figure}

In addition to the NER model, a regular expressions module is used to detect other sensitive entities, namely, Spanish ID number (DNI/NIE), email address, bank account numbers, phone number in both landline or mobile pattern formats and Spanish health card number. The patterns used to find these pieces of information are obtained through data reviewing. Once any of these entities is detected, the substitution module replaces it with a fake (random) but realistic string of characters, resembling the entity that needs to be replaced (see Figure \ref{iban_substitution}). Realistic but fake entity replacement is applied to maintain the consistency and coherence of the contexts in which the entities appear.

A cyclical qualitative peer review is conducted throughout the development of the anonymization tool and after the anonymization process to ensure accuracy and minimize the exposure of sensitive data. After each revision, adjustments to the anonymization code are made to improve results without affecting the quality of previously obtained results. This quality review is key for the corpus to be further annotated and used in fine-tuning.

\begin{figure}
\centering
\includegraphics[width=0.9\textwidth]{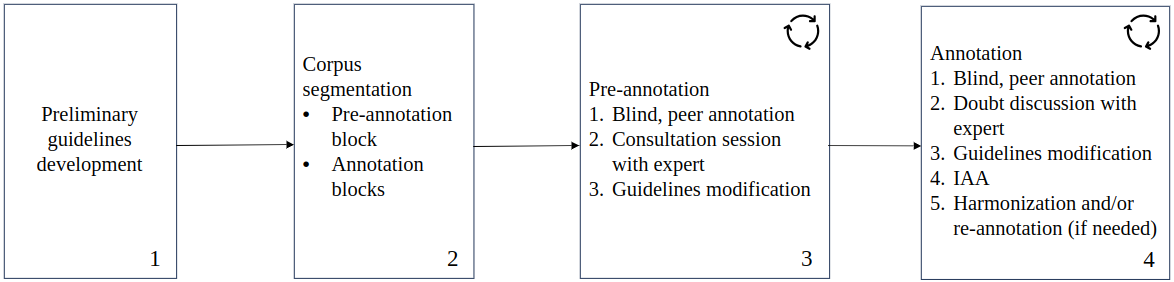}
    \caption{Annotation process scheme.}
    \label{annotation_process_scheme}
\end{figure}

\textbf{Annotation methodology.} The annotation process followed the principles of MATTER methodology \cite{pustejovskystubbs12},\cite{aldamaetal22} which covers four main phases: developing preliminary annotation guides, corpus segmentation, pre-annotation and annotation (see Figure \ref{annotation_process_scheme}). 

The first step in the annotation process is to create preliminary annotation guides that aim to cover basic label definitions. Preliminary guidelines are taken as the basis to perform pre-annotation. To develop preliminary guidelines, an expert in private and public contracts was involved to clarify legal term definitions and adjust the scope of each of the labels. The kind of annotation chosen to account for the proposed tag set is span categorization, which consists in selecting strings of text (sentences in the case of 3CEL) that match label definitions.    

The second step is related to corpus segmentation. The filtered, clean and anonymized version of the data is randomly divided into 19 blocks of \textasciitilde{~}20 contracts each\footnote{The pre-annotation block is not included in the 3CEL. The first 18 annotation blocks contain 20 contracts. The last one contains 13 contracts.}. One of the blocks is used in the pre-annotation phase. The rest of the blocks are annotated reiteratively during the annotation phase. 

Pre-annotation follows corpus segmentation. The pre-annotation phase is conducted to test and tune the applicability of preliminary guides over real data (the pre-annotation block mentioned above).  At this point it is important to mention that both pre-annotation and annotation processes are carried out in a blind and peer-reviewed way, meaning that two or more linguists work on the same texts at the same time, using the annotation guidelines as the only means of information. Annotation guides serve as the foundational tool to provide instructions on how to label the data, ensuring consistency and precision across annotators. Thus, the guides are modified and detailed during both the pre-annotation and annotation phases. Blind, peer-reviewed annotation is key to ensure consistency and avoid personal biases. Once pre-annotation concluded, annotators shared and discussed ambiguities and doubts, that were solved by the contract expert. Annotation guidelines were modified to account for the discussed cases. 

\begin{figure}[h]
\centering
\includegraphics[width=1\textwidth]{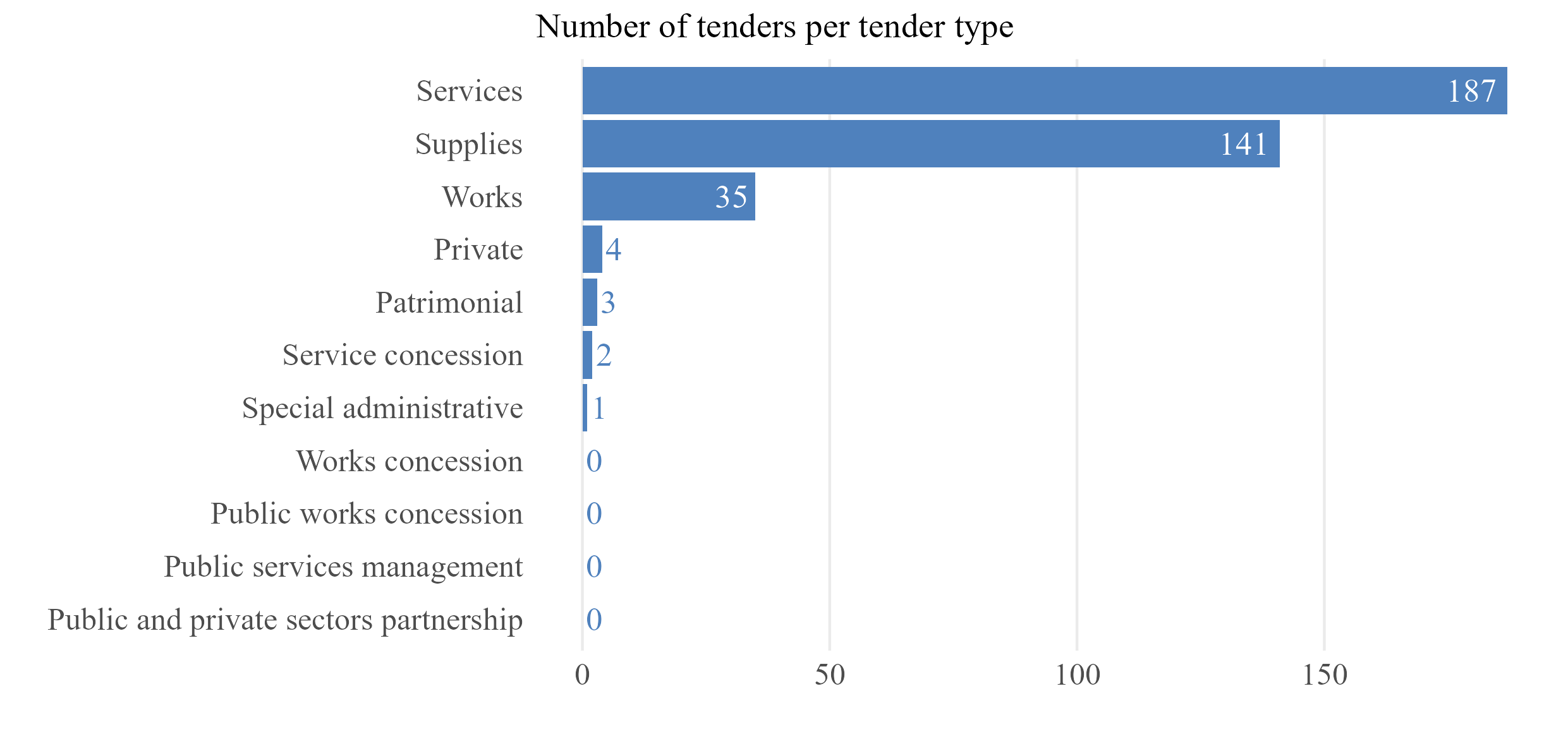}
    \caption{3CEL tender typology distribution.}
    \label{3CEL_tender_typology}
\end{figure}

\begin{wraptable}{r}{8cm}
  \centering
  \begin{tabular}{lr}
    \toprule
    \multicolumn{2}{c}{\textbf{3CEL}}   \\
    \midrule
    Number of contracts                 & 373       \\
    Annotation blocks                   & 19        \\
    Number of assigned tags             & 4782      \\
    Average number of tags per contract & 12.8      \\
    Categories defined in annotation guides & 19
    \\
    Categories present in corpus        & 18 
    \\
    Average IAA                         & 0.61      \\
    \bottomrule \\
  \end{tabular}
    \caption{Summary of 3CEL features}
    \label{tab:corpus_features}
\end{wraptable}

\begin{figure}
\centering
\includegraphics[width=1\textwidth]{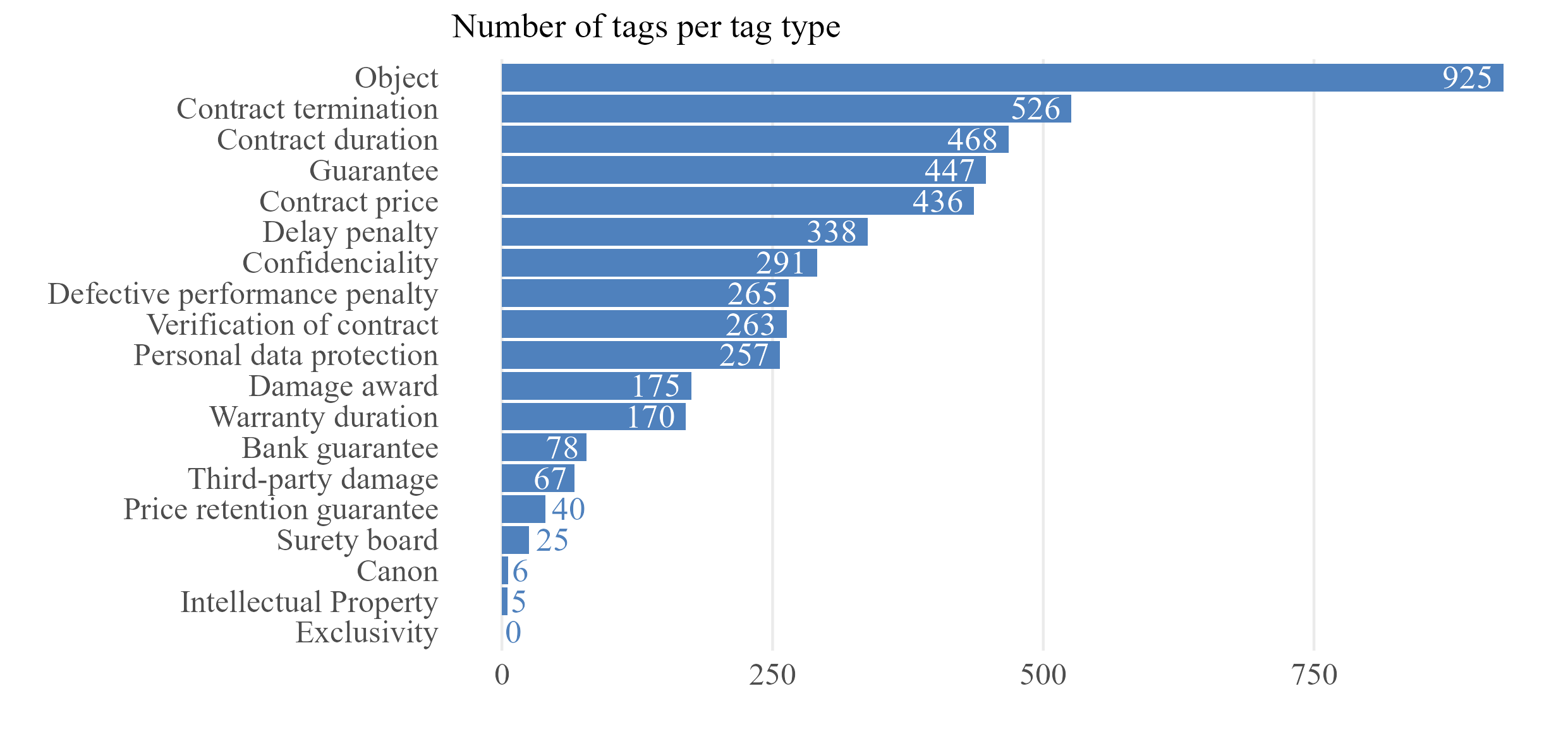}
    \caption{3CEL tag distribution.}
    \label{3CEL_tag_distribution}
\end{figure}

Lastly, annotation is performed reiteratively as described in Figure \ref{annotation_process_scheme} using Prodigy \cite{prodigy18}. For each of the 19 annotation blocks, two annotators simultaneously tag tenders applying the guidelines criteria, independently collect doubts and ambiguities, that are discussed with the legal expert and resolved. The resolution of the discussed questions is included in the guidelines. Inter-annotator agreement (IAA) metric is calculated to assess the level of agreement between both annotators. The chosen standard for IAA evaluation is the strictest: annotators must match both the category and the annotation span. For this project, the mean IAA score is 0.61. After IAA calculation, a harmonization process takes part to resolve discrepancies and align annotations. Since the tag set and criteria varied through annotation, re-annotation was required in certain blocks to maintain tagging homogeneity across the entire annotated corpus. Table \ref{tab:corpus_features} and Figures \ref{3CEL_tender_typology} and \ref{3CEL_tag_distribution} summarize global corpus information and tag distribution. 

\section{Contract clause fine-tuning} 
This section presents a set of experiments carried out with different models and configurations using 3CEL to solve clause extraction and categorization in Spanish tenders.

\subsection{Setup}
\textbf{Task structure.} The goal of a span categorization fine-tuning task is to make the model learn to tag the text spans that belong to a certain category. This is a hybrid between a multilabel classification and NER problem. In this regard, span categorization models face three challenges: spans length, multi-labels and label-doubling. First, in NER cases, the potential length an entity may display is no longer than a sentence, whereas in span categorization, a span may be several sentences long. Second, a typical NER task implies multi-class classification, and thus, one single tag is associated to a particular span. Meanwhile, span categorization allows associating multiple tags to one single span. Furthermore, the spans annotated are divided into beginning (B) and inner (I) depending on the position a particular token occupies within a span (e.g. B-OBJECT vs. I-OBJECT). Thus, the number of annotated tags and the number of labels in the tag set are doubled, including 36 classification tags instead of the original 18. These three issues make span categorization a complex task to solve.

\textbf{Metrics.} Macro F1-score is chosen to measure model performance because categories are imbalanced (see Figure \ref{3CEL_tag_distribution}), but equally significant for the purposes of this IE task. Macro F1-score is the arithmetic mean of all the F1 scores per class. F1-score is the harmonic mean of precision and recall. Precision is the proportion of all relevant retrieved pieces of information out of the total amount of retrieved information. Recall is the proportion of relevant retrieved pieces of information out of all the relevant pieces of information. 

\textbf{Models.} The perfomance of \texttt{xlm-roberta-large} \cite{xlm-roberta-large19}, \texttt{legal-xlm-roberta-large} \cite{niklaus23}, \texttt{RoBERTalex} \cite{gutierrezfandino21} and \texttt{MEL} \cite{betancur25} \footnote{MEL is also developed in the context of INESData.} is evaluated. \texttt{xlm-roberta-large} is taken as baseline.

\textbf{Fine-tuning.} 3CEL is processed to be randomly divided in train (75\,\% of full tenders) and test (25\,\% of full tenders) datasets. Train and test datasets are maintained through the set of experiments for the sake of result comparability. Train and test datasets are separately split into chunks smaller than 512 tokens to fit inside the models contextual windows, preserving entities within the margins of each of the chunks. The train and test datasets contain 1\,706 and 604 chunks, respectively. 
\newpage
\begin{wraptable}{r}{8cm}
  \centering
  \begin{tabular}{ll}
    \toprule
    \multicolumn{2}{c}{\textbf{Fine-tuning parameters}}   \\
    \midrule
    evaluation\_strategy  & epoch  \\
    logging\_strategy     & epoch  \\
    learning\_rate        & 2e-5 \\
    per\_device\_train\_batch\_size & 2\\
    per\_device\_eval\_batch\_size  & 2\\
    gradient\_accumulation\_steps   & 2 \\
    num\_train\_epochs    & 30\\
    weight\_decay         & 0.01\\
    save\_strategy        & epoch\\
    save\_total\_limit    & 2\\
    load\_best\_model\_at\_end & True\\
    metric\_for\_best\_Bearing in mindmodel  & macro\_f1 \\
    seed                       & 12345\\
    \bottomrule \\
  \end{tabular}
    \caption{Summary of span categorization fine-tuning parameters}
    \label{tuning_parameters}
\end{wraptable}
The experiments are conducted in \texttt{g4dn.4xlarge} AWS instances. The span categorization fine-tuning code used is based on \cite{tutorialspancat22} and the HuggingFace Transformers library \cite{wolfetal00}. After fine-tuning, models must predict the spans where a concrete tag applies assigning a confidence probability rate. The models mentioned above are fine-tuned using the whole set of labels (18 categories) and a reduced set of labels (15 categories) with the largest representation (over 40 appearances). 
Models are fine-tuned according to the parameters in Table \ref{tuning_parameters}. All the parameters display default values, except for gradient\_accumulation\_steps. Due to processing limitations, the batch size could not be greater than 2. Consequently, the parameter gradient\_accumulation\_steps is set to 2, so the model goes through 4 text segments before updating its weights. Additionally, the training dataset is shuffled before fine-tuning to minimize the chances for the model to adjust its weights on the basis of 4 text segments coming from the same contract. The optimizer and network classification head layers are those set by default by Transformers 4.42.4 version.

\subsection{Results} 
\begin{figure}
  \begin{minipage}[b]{0.49\textwidth}
    \includegraphics[width=\textwidth]{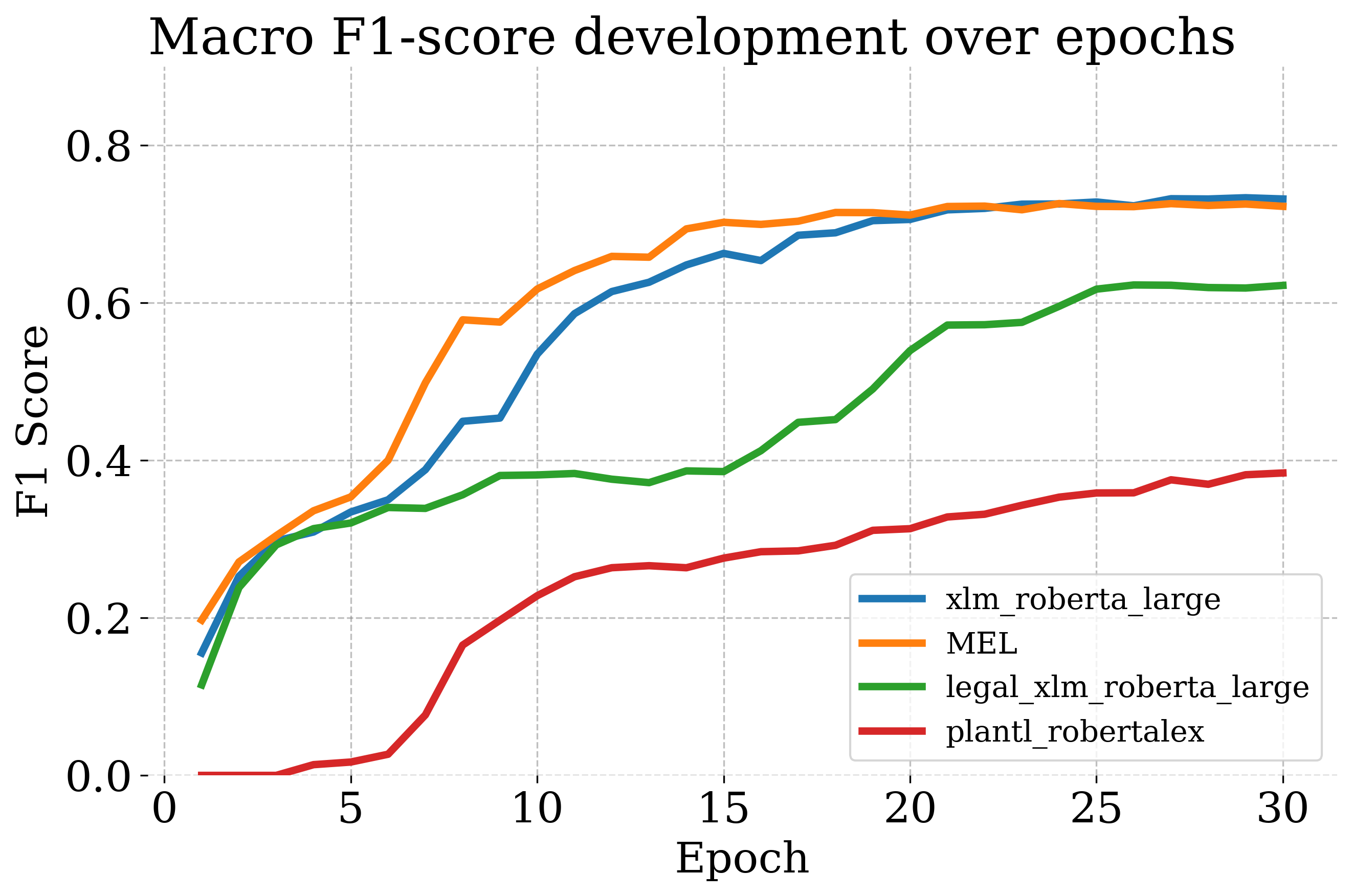}
    \caption{F1-score development over fine-tuning with 18 categories}
    \label{f1score_18cats}
  \end{minipage}
  \hfill
  \begin{minipage}[b]{0.49\textwidth}
    \includegraphics[width=\textwidth]{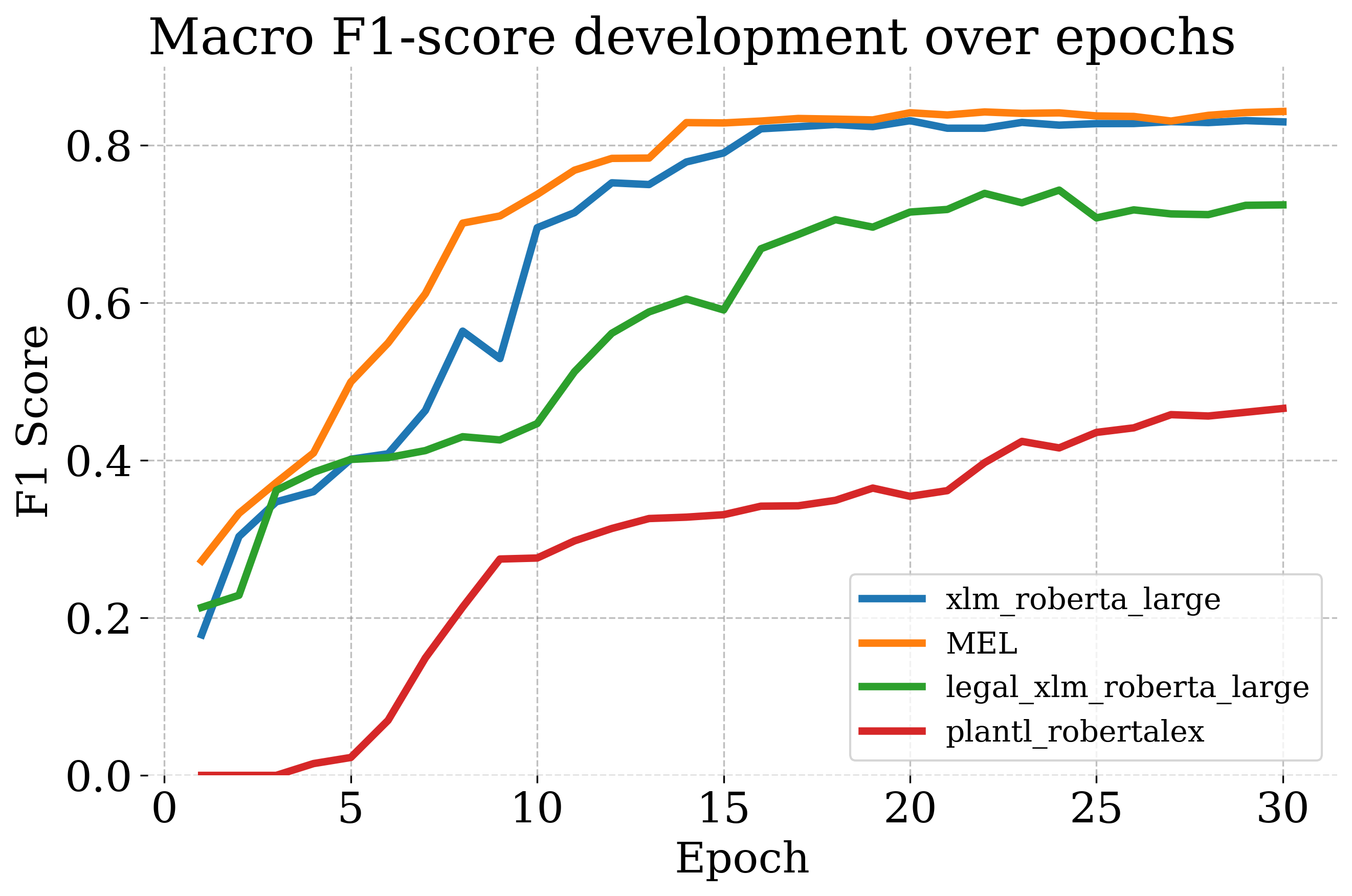}
    \caption{F1-score development over fine-tuning with 15 categories}
    \label{f1score_15cats}
  \end{minipage}
\end{figure}

\begin{figure}
  \begin{minipage}[b]{0.49\textwidth}
    \includegraphics[width=\textwidth]{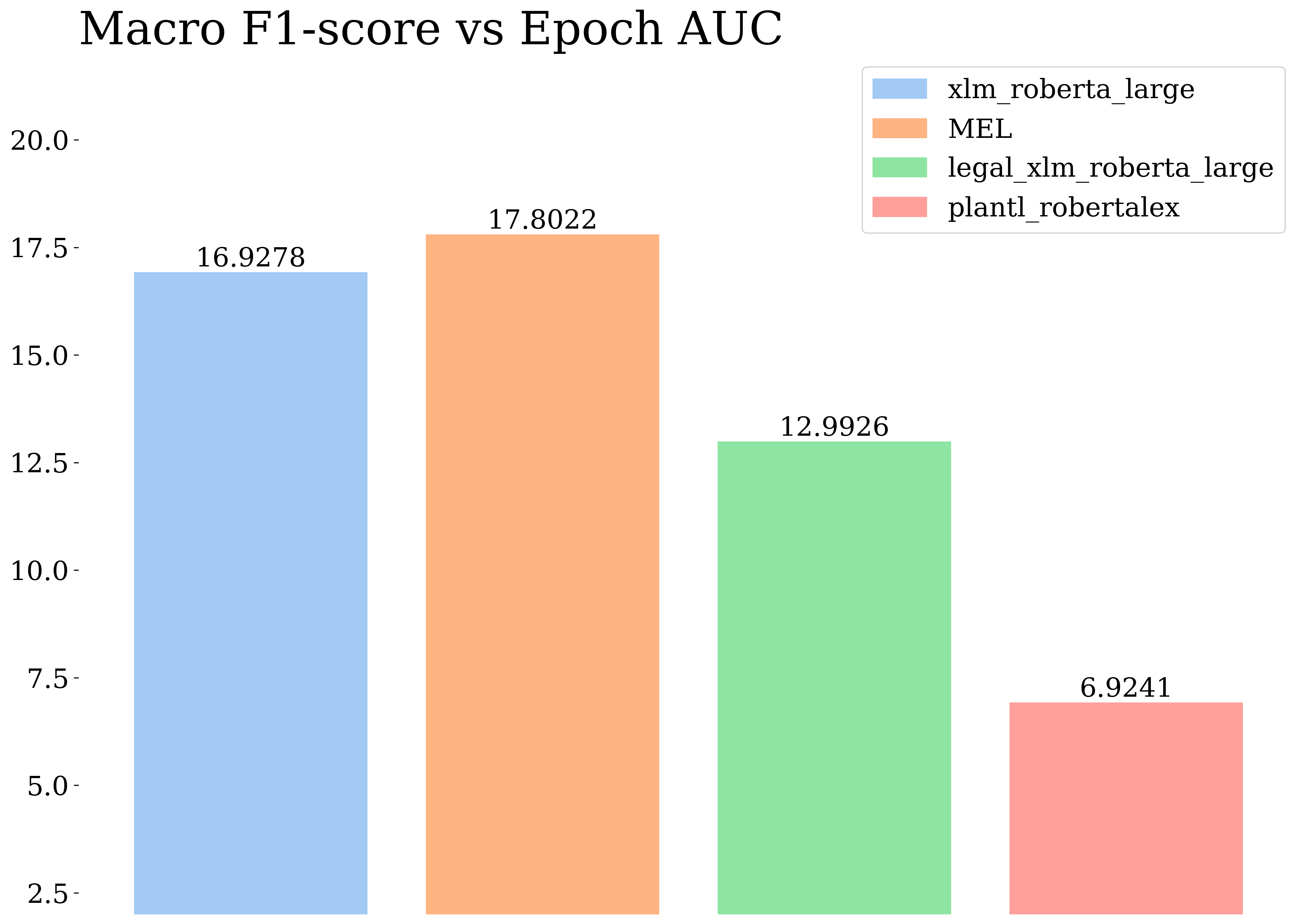}
    \caption{Epochs vs macro F1-score AUC (18 categories)}
    \label{auc_18cats}
  \end{minipage}
  \hfill
  \begin{minipage}[b]{0.49\textwidth}
    \includegraphics[width=\textwidth]{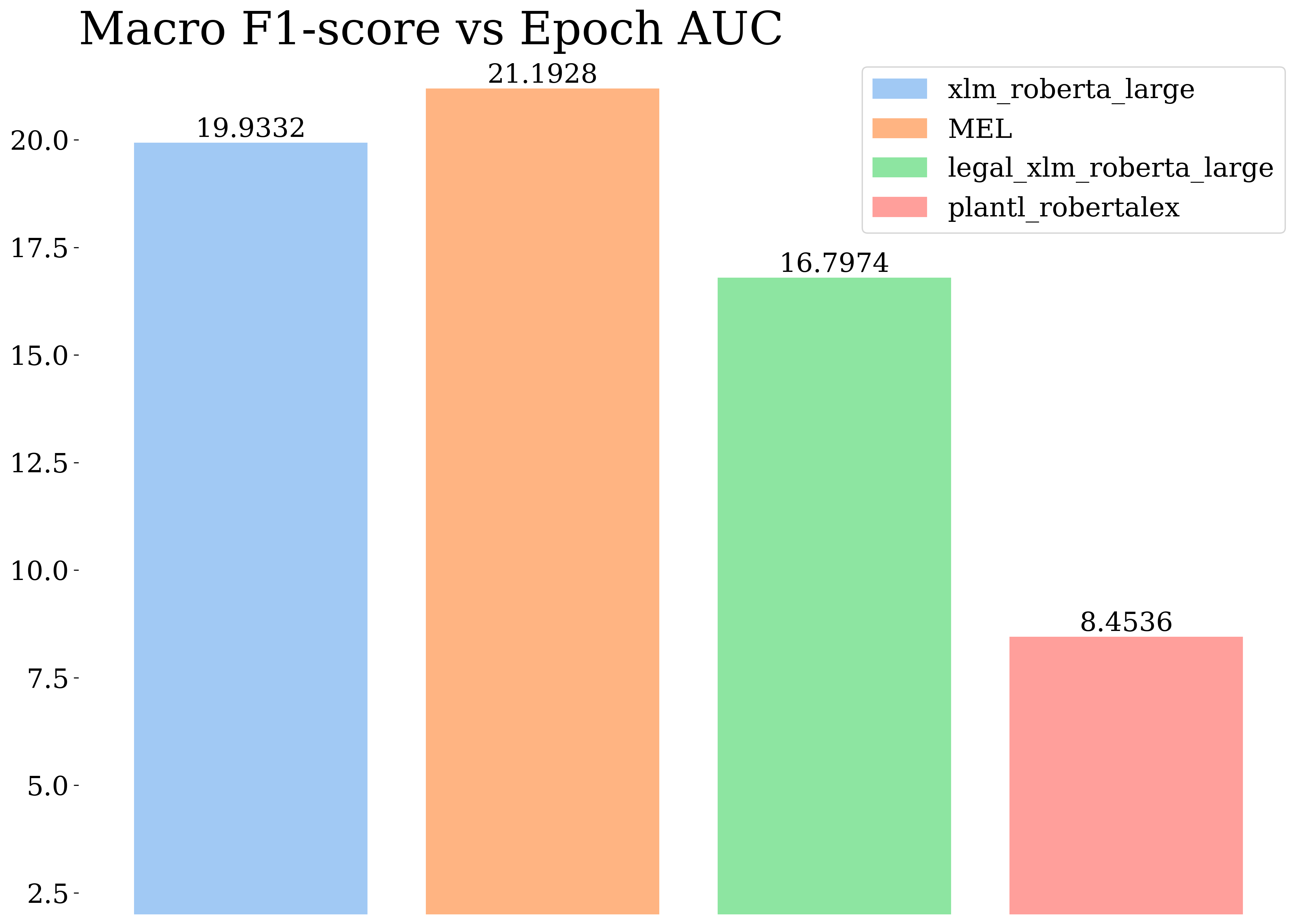}
    \caption{Epochs vs macro F1-score AUC (15 categories)}
    \label{auc_15cats}
  \end{minipage}
\end{figure}

Table \ref{model_results} shows the training loss, validation loss and macro F1-score values obtained after fine-tuning \texttt{xlm-roberta-large}, \texttt{legal-xlm-roberta-large}, \texttt{RoBERTalex} and \texttt{MEL} in the same conditions. Figures \ref{f1score_18cats} and \ref{f1score_15cats} show the development of macro F1-scores over fine-tuning with 18 and 15 labels, respectively. That is, the degree of performance the models achieve regarding macro F1-score for each of the training epochs. Figures \ref{auc_18cats} and \ref{auc_15cats} show the models performance in terms of learning speed, representing epochs versus macro F1-score areas under the curve (AUC). These figures represent the models capacity to obtain better results in early stages of the fine-tuning process, in other words, the larger the area, the greater the learning degree of the model over training.

 \begin{table}[h]
    \begin{center}
    \begin{tabular}{llrrr}
        \toprule
        \textbf{Categories} & \textbf{Model} & \textbf{Training loss} & \textbf{Validation loss} & \textbf{Macro F1-score}\\
        \midrule
        \multirow{4}{7em}{18 labels}
        & xlm-roberta-large & 0.0002 & 0.0095 & \textbf{0.73} \\ 
        & legal-xlm-roberta-large & 0.0002 & 0.0077 & 0.62\\ 
        & RoBERTalex & 0.0021 & 0.0085 & 0.38\\
        & MEL & 0.0001 & 0.0104
 & 0.72\\
        \midrule
        \multirow{4}{7em}{15 labels > 40 appearances}
        & xlm-roberta-large & 0.0002 & 0.0105 & 0.83\\ 
        & legal-xlm-roberta-large & 0.0001 & 0.0084 & 0.72\\ 
        & RoBERTalex & 0.0021 & 0.0098 & 0.47\\
        & MEL & 0.0002 & 0.0111
 & \textbf{0.84}\\
        \bottomrule
    \end{tabular}
        \caption{Results of NLP models on 3CEL}
        \label{model_results}
    \end{center}
\end{table}

Regarding the 18-label set of experiments, \texttt{xlm-roberta-large} is the model that best performs ($\sim$0.73 macro F1-score), followed by MEL ($\sim$0.72 macro F1-score). This small difference is not relevant since \texttt{MEL}, which is based on \texttt{xlm-roberta-large}, is trained on more data than \texttt{xlm-roberta-large}, and that implies more trustful results. In addition, as shown by Figures \ref{f1score_18cats} and \ref{auc_18cats}, \texttt{MEL} is the fastest learning model: \texttt{MEL} consistently obtains better F1-score values than xlm-roberta-large until epoch 20, when both models converge.

Regarding the 15 labels set of experiments, \texttt{MEL} outperforms the rest of models ($\sim$0.84 macro F1-score), followed by \texttt{xlm-roberta-large} ($\sim$0.82 macro F1-score). Again, Figures \ref{f1score_15cats} and \ref{auc_15cats} show that \texttt{MEL} is the fastest learning model in this 15-label scenario, obtaining the greatest epochs vs macro F1-score AUC.
Ruling out the three less represented categories (canon, surety board and intellectual property) implies an improvement of $\sim$0.1 in performance. 

Considering that \texttt{MEL} is the fastest learning model and the one trained on the greatest amount of data out of the four models, \texttt{MEL} is concluded to be the best model of the experimental set.

\section{Conclusions}
This paper introduces 3CEL, which is a high quality resource to explore information extraction in the Spanish legal domain. 3CEL contains 373 tenders that are manually annotated using 19 
 defined categories (4\,782 total tags) that represent interesting information for contract understanding and reviewing. 3CEL is used to fine-tune different state-of-the-art models with 18 out of the 19 labels and the 15 most represented ones. Among these models \texttt{MEL} obtains the second best and best results in macro F1-score with 18 and 15 tags, respectively, and it is the fastest learning model of the set of experiments. The experiments carried out validate 3CEL as a valuable resource for span categorization.
Further work includes expanding 3CEL to include private sector contracts and explore modeling results with coming models and architectures.

\textbf{Acknowledgements}

This work has received funding from the INESData project (Infrastructure to Investigate Data Spaces in Distributed Environments at UPM), a project funded under the UNICO I+D CLOUD call by the Ministry for Digital Transformation and the Civil Service, in the framework of the recovery plan PRTR financed by the European Union (NextGenerationEU).
The completion of this work would not have been possible without the help and dedication of Maite Sanz de Galdeano, Borja Adsuara Varela and Alfonso Egea de Haro.

\bibliographystyle{unsrt}  
\bibliography{references}

\begin{thebibliography}{10}

\bibitem{grishman97}
Ralph Gishman.
\newblock Information extraction: Techniques and challenges.
\newblock In {\em SCIE '97: International Summer School on Information Extraction: A Multidisciplinary Approach to an Emerging Information Technology}, pages 10--27. SCIE, 1997.

\bibitem{piskorskiyangarber12}
Jakub Piskorski and Roman Yangarber.
\newblock {\em Information Extraction: Past, Present and Future}, pages 23--49.
\newblock Springer, United States, 2012.

\bibitem{dagdelenetal24}
John Dagdelen, Alexander Dunn, Sanghoon Lee, Nicholas Walker, Andrew~S. Rosen, Gerbrand Ceder, Kristin~A. Persson, and Anubhav Jain.
\newblock Structured information extraction from scientific text with large language models.
\newblock {\em Nature Communications}, 2024.

\bibitem{Legalruleml21}
Monica Palmirani, Guido Governatori, Tara Athan, Harold Boley, Adrian Paschke, and Adam Wyner.
\newblock Legalruleml core specification version 1.0.
\newblock \url{https://docs.oasis-open.org/legalruleml/legalruleml-core-spec/v1.0/legalruleml-core-spec-v1.0.html}, 2021.

\bibitem{fibo99}
Enterprise Data Management Council~EDM Council.
\newblock Financial industry business ontology - fibo.
\newblock \url{https://edmcouncil.org/frameworks/industry-models/fibo/}, 1999--2024.

\bibitem{contractonto20}
CSM Lab~School of~Electrical~Engineering and Computer Science (EECS)~University of~Ottawa.
\newblock Contract ontology.
\newblock \url{https://sites.google.com/uottawa.ca/csmlab/research/contract-ontology}, 2020.

\bibitem{martinchozas21}
Patricia Martín-Chozas and Artem Revenko.
\newblock Thesaurus enhanced extraction of hohfeld's relations from {S}panish labour law.
\newblock In Sarra~Ben Abb{\`{e}}s, Rim Hantach, Philippe Calvez, Davide Buscaldi, Danilo Dess{\`{\i}}, Mauro Dragoni, Diego~Reforgiato Recupero, and Harald Sack, editors, {\em Joint Proceedings of the 2nd International Workshop on Deep Learning meets Ontologies and Natural Language Processing (DeepOntoNLP 2021) {\&} 6th International Workshop on Explainable Sentiment Mining and Emotion Detection {(X-SENTIMENT} 2021) co-located with co-located with 18th Extended Semantic Web Conference 2021, Hersonissos, Greece, June 6th - 7th, 2021 (moved online)}, volume 2918 of {\em {CEUR} Workshop Proceedings}, pages 30--38. CEUR-WS.org, 2021.

\bibitem{prooc16}
José Muñoz-Soro, Guillermo Esteban, Oscar Corcho, and Francisco Serón.
\newblock Pproc, an ontology for transparency in public procurement.
\newblock {\em Semantic Web}, 7:295--309, 03 2016.

\bibitem{theybuyforyouxx}
They buy for you.
\newblock \url{https://theybuyforyou.eu}.

\bibitem{nextprocurement}
Barcelona supercomputing Center, Universidad Carlos~III de~Madrid, Universidad~Politécnica de~Madrid, LocaliData, Generalitat de~Catalunya, Centre de~Telecomunicacions i Tecnologies de~la informació, Ayuntamiento de~Madrid, and Ayuntamiento de~Zaragoza.
\newblock Nextprocurement.
\newblock \url{http://nextprocurement-project.com}.

\bibitem{montiel17}
Elena Montiel-Ponsoda, Víctor Rodríguez-Doncel, and Jorge Gracia.
\newblock Building the legal knowledge graph for smart compliance services in multilingual europe.
\newblock In {\em In Proceedings of the 1st Workshop on Technologies for Regulatory Compliance.}, 2017.

\bibitem{martinchozas19}
Patricia Martín-Chozas, Elena Montiel-Ponsoda, and Víctor Rodríguez-Doncel.
\newblock {\em Language Resources as Linked Data for the Legal Domain}, pages 170--180.
\newblock IOS Press BV, Amsterdam, 2019.

\bibitem{atticus23}
The~Atticus Project.
\newblock The atticus project.
\newblock \url{https://www.atticusprojectai.org}, 2023.

\bibitem{cuad21}
Dan Hendrycks, Collin Burns, Anya Chen, and Anya Ball.
\newblock {CUAD:} an expert-annotated {NLP} dataset for legal contract review.
\newblock {\em CoRR}, abs/2103.06268, 2021.

\bibitem{BERT}
Jacob Devlin, Ming-Wei Chang, Kenton Lee, and Kristina Toutanova.
\newblock Bert: Pre-training of deep bidirectional transformers for language understanding, 2019.

\bibitem{ALBERT}
Zhenzhong Lan, Mingda Chen, Sebastian Goodman, Kevin Gimpel, Piyush Sharma, and Radu Soricut.
\newblock Albert: A lite bert for self-supervised learning of language representations, 2020.

\bibitem{ROBERTA}
Yinhan Liu, Myle Ott, Naman Goyal, Jingfei Du, Mandar Joshi, Danqi Chen, Omer Levy, Mike Lewis, Luke Zettlemoyer, and Veselin Stoyanov.
\newblock Roberta: A robustly optimized bert pretraining approach, 2019.

\bibitem{DEBERTA}
Pengcheng He, Xiaodong Liu, Jianfeng Gao, and Weizhu Chen.
\newblock Deberta: Decoding-enhanced bert with disentangled attention, 2021.

\bibitem{gutierrezfandino2021legal}
Asier Gutiérrez-Fandiño, Jordi Armengol-Estapé, Aitor Gonzalez-Agirre, and Marta Villegas.
\newblock Spanish legalese language model and corpora, 2021.

\bibitem{niklaus23}
Joel Niklaus, Veton Matoshi, Matthias Sturmer, Ilias Chalkidis, and Daniel~E. Ho.
\newblock Multilegalpile: A 689gb multilingual legal corpus.
\newblock {\em ArXiv}, abs/2306.02069, 2023.

\bibitem{littlejohn24}
Little~John AI.
\newblock bge-m3-spa-law-qa.
\newblock https://huggingface.co/littlejohn-ai/bge-m3-spa-law-qa, 2024.

\bibitem{aldamaetal22}
Nuria Aldama, Marta Guerrero, Helena Montoro, and Doaa Samy.
\newblock Anotación de corpus lingüísticos: metodología utilizada en el instituto de ingeniería del conocimiento (iic).
\newblock \url{https://www.iic.uam.es/whitepapers/anotacion-corpus-linguisticos-metodologia-utilizada-iic/}, 2022.

\bibitem{contratacionpublica08}
Dirección~General del Patrimonio del Estado del Ministerio de Hacienda~y Administraciones~Públicas.
\newblock Plataforma de contratación del sector público.
\newblock \url{https://contrataciondelestado.es/wps/portal/plataforma}, 2008.

\bibitem{ocrmypdf23}
James~R. Barlow.
\newblock Ocrmypdf.
\newblock \url{https://ocrmypdf.readthedocs.io/en/latest/}, 2023.

\bibitem{pdftotext21}
Jason~Alan Palmer.
\newblock pdftotext 2.2.2.
\newblock \url{https://pypi.org/project/pdftotext/}, 2021.

\bibitem{pdfplumber24}
Jeremy Singer-Vine.
\newblock pdfplumber 0.11.4.
\newblock \url{https://pypi.org/project/pdfplumber/}, 2024.

\bibitem{pypdf222}
Mathieu Fenniak.
\newblock Pypdf2 3.0.1.
\newblock \url{https://pypi.org/project/PyPDF2/}, 2022.

\bibitem{mmg23}
dezzai.
\newblock Mmg/xlm-roberta-large-ner-spanish.
\newblock \url{https://huggingface.co/MMG/xlm-roberta-large-ner-spanish}, 2023.

\bibitem{pustejovskystubbs12}
James Pustejovsky and Amber Stubbs.
\newblock {\em Natural Language Annotation for Machine Learning}.
\newblock O'Reilly Media, Inc., 2012.

\bibitem{prodigy18}
Ines Montani and Matthew Honnibal.
\newblock Prodigy: A new annotation tool for radically efficient machine teaching.
\newblock {\em Artificial Intelligence}, 2018.

\bibitem{xlm-roberta-large19}
Alexis Conneau, Kartikay Khandelwal, Naman Goyal, Vishrav Chaudhary, Guillaume Wenzek, Francisco Guzm{\'{a}}n, Edouard Grave, Myle Ott, Luke Zettlemoyer, and Veselin Stoyanov.
\newblock Unsupervised cross-lingual representation learning at scale.
\newblock {\em CoRR}, abs/1911.02116, 2019.

\bibitem{gutierrezfandino21}
Asier Gutiérrez-Fandiño, Jordi Armengol-Estapé, Aitor Gonzalez-Agirre, and Marta Villegas.
\newblock Spanish legalese language model and corpora, 2021.

\bibitem{betancur25}
David Betancur~Sánchez, Nuria Aldama~García, Álvaro Barbero~Jiménez, Marta Guerrero~Nieto, Patricia Marsà~Morales, Nicolás Serrano~Salas, Carlos García~Hernán, Pablo Haya~Coll, Elena Montiel~Ponsoda, and Pablo Calleja~Ibáñez.
\newblock Mel: Legal {S}panish {L}anguage {M}odel.
\newblock {\em ArXiv}, To be published.

\bibitem{tutorialspancat22}
La~Javaness R\&D.
\newblock Finetune a span categorizer with bert and transformers.
\newblock \url{https://lajavaness.medium.com/1-token-classification-vs-span-categorization-52a685e4674a}, 2022.

\bibitem{wolfetal00}
Thomas Wolf, Lysandre Debut, Victor Sanh, Julien Chaumond, Clement Delangue, Anthony Moi, Pierric Cistac, Tim Rault, Rémi Louf, Morgan Funtowicz, Joe Davison, Sam Shleifer, Patrick von Platen, Clara Ma, Yacine Jernite, Julien Plu, Canwen Xu, Teven Le~Scao, Sylvain Gugger, Mariama Drame, Quentin Lhoest, and Alexander~M. Rush.
\newblock Transformers: State-of-the-art natural language processing.
\newblock In {\em In Proceedings of the 2020 Conference on Empirical Methods in Natural Language Processing: System Demonstrations. Association for Computational Linguistics, 2020.}, 2000.

\end{thebibliography}

\end{document}